\PassOptionsToPackage{table,xcdraw,dvipsnames}{xcolor}

\documentclass{article}

\PassOptionsToPackage{numbers, compress}{natbib}

\usepackage[preprint]{neurips_2025}

\usepackage[utf8]{inputenc} 
\usepackage[T1]{fontenc}    
\usepackage{hyperref}       
\hypersetup{
    colorlinks=true,
    linkcolor=BlueViolet,
    citecolor=teal
}
\usepackage{url}            
\usepackage{booktabs}       
\usepackage{amsfonts}       
\usepackage{nicefrac}       
\usepackage{microtype}      
\usepackage{xcolor}         
\usepackage{ulem}
\usepackage{makecell}
\usepackage{amsthm}
\usepackage{bm}
\usepackage[table,xcdraw]{xcolor}
\usepackage{multicol}
\usepackage{colortbl}
\usepackage{amsmath}
\usepackage{cleveref}
\usepackage{multirow}
\usepackage{graphicx}
\usepackage{subcaption}
\usepackage{enumitem}
\usepackage{bbding}
\usepackage{color}

\usepackage{algorithm}
\usepackage{listings}
\usepackage{natbib}
\usepackage{tcolorbox}

\usepackage{mdframed}
\usepackage{amsmath,bm,bbm}
\usepackage{footnote}
\usepackage{adjustbox}

\usepackage{epigraph}
\definecolor{baselinecolor}{gray}{.9}

\usepackage{etoolbox}
\makeatletter
\AfterEndEnvironment{algorithm}{\let\@algcomment\relax}
\AtEndEnvironment{algorithm}{\kern2pt\hrule\relax\vskip3pt\@algcomment}
\let\@algcomment\relax
\newcommand\algcomment[1]{\def\@algcomment{\footnotesize#1}}
\renewcommand\fs@ruled{\def\@fs@cfont{\bfseries}\let\@fs@capt\floatc@ruled
  \def\@fs@pre{\hrule height.8pt depth0pt \kern2pt}%
  \def\@fs@post{}%
  \def\@fs@mid{\kern2pt\hrule\kern2pt}%
  \let\@fs@iftopcapt\iftrue}
\makeatother
\definecolor{tabhighlight}{HTML}{e5e5e5}
\definecolor{tabhighlight2}{HTML}{e8e8e8}

\newcommand{\para}[1]{
  \noindent\textbf{#1}
}
\newcommand{\name}{CompassJudger-2}
\newcommand{\model}[1]{CompassJudger-2-{#1}-Instruct}
\newcommand{\red}[1]{\textcolor{red}{#1}}
\usepackage{listings}
\definecolor{ForestGreen}{RGB}{34,139,34}

\renewcommand{\paragraph}[1]{\medskip\noindent\textbf{#1.~}}

\theoremstyle{plain}
\newmdtheoremenv{corollary}{Corollary}

\newmdtheoremenv[linewidth=0pt,innerleftmargin=4pt,innerrightmargin=4pt]{lemma}{Lemma}%

\usepackage{tikz}

\newcommand*{\circled}[1]{\lower.7ex\hbox{\tikz\draw (0pt, 0pt)%
    circle (.5em) node {\makebox[1em][c]{\small #1}};}}

\title{CompassJudger-2: Towards Generalist Judge Model \\ via Verifiable Rewards}
\author{Taolin Zhang$^{1,2,*}$, Maosong Cao$^{1,*}$, Alexander Lam$^{1}$, Songyang Zhang$^{1,\dagger,\ddagger}$, Kai Chen$^{1,\dagger}$\\
$^1$Shanghai AI Laboratory $^2$Tsinghua University\\
\texttt{Github:\href{https://github.com/open-compass/CompassJudger}{https://github.com/open-compass/CompassJudger}}\\
}

\begin{document}

\maketitle
\begin{abstract}

Recently, the role of LLM-as-judge in evaluating large language models has gained prominence. However, current judge models suffer from narrow specialization and limited robustness, undermining their capacity for comprehensive  evaluations. In this work, we present \textit{CompassJudger-2}, a novel generalist judge model that overcomes these limitations via a task-driven, multi-domain data curation strategy. Central to our approach is supervising judgment tasks with verifiable rewards, guiding intrinsic critical reasoning through rejection sampling to foster robust, generalizable judgment capabilities. We introduce a refined learning objective with margin policy gradient loss to enhance performance. Empirically, CompassJudger-2 achieves superior results across multiple judge and reward benchmarks, and our 7B model demonstrates competitive judgment accuracy with significantly larger models like DeepSeek-V3 and Qwen3-235B-A22B. Additionally, we propose JudgerBenchV2, a comprehensive benchmark evaluating cross-domain judgment accuracy and rank consistency to standardize judge model evaluation. These contributions advance robust, scalable LLM judgment and establish new performance and evaluation standards. \footnote{This work is done when Taolin Zhang 
is on internship at Shanghai AI Laboratory, * means equal
contribution, $^{\dagger}$ means corresponding author, $^{\ddagger}$ means project
lead.}

\end{abstract}
\section{Introduction}
\label{sec: introduction}

In recent years, large language models (LLMs) have advanced rapidly with the development of new foundation models such as DeepSeek-R1~\cite{guo2025deepseek}, OpenAI-o1~\cite{jaech2024openai}, and the Qwen series~\cite{yang2024qwen2}. Innovations in architecture and data scaling have enabled LLMs to achieve state-of-the-art performance across diverse tasks, including natural language understanding, code generation, creative writing, and complex reasoning~\cite{li2024crowdsourced, jain2024livecodebench, dubois2024length, lambert2024rewardbench}.

As LLMs are deployed in real-world applications, accurate evaluation of response quality has become increasingly critical. Rule-based benchmarks ~\cite{rein2024gpqa,zhou2023instruction,jain2024livecodebench, clark2019boolqexploringsurprisingdifficulty, 2023opencompass, dua2019dropreadingcomprehensionbenchmark} excel at evaluating standardized tasks but struggle with LLM output variability, often failing to handle edge cases due to reliance on complex regex designs. Model-based approaches like Reward Models and LLM-as-Judge ~\cite{zhu2023judgelm,chen2025xverify,ye2024scalarrewardmodellearning,yu2025improve} reduce evaluation efforts by leveraging the reasoning ability of LLMs. However, these approaches introduce some new challenges in that restricted generalization ability of existing judge models confines them to specific prompts or datasets. Moreover, some inadequate world knowledge of these LLMs may lead to inaccurate judgments on knowledge-intensive queries, limiting their application for iterative model improvement. 


To address these limitations, we propose a unified training paradigm for judge models. First, we define a series of potential application scenarios for judge models and collect a wide range of judge-related public datasets. Subsequently, we curate and synthesize data from different sources to obtain a diverse training dataset. Second, we employ judgment-oriented chain-of-thought (CoT) data generation to improve judgment accuracy, combined with rejection sampling to select high-quality training examples. Finally, we introduce a margin policy gradient loss with verifiable reward signals for better optimization. The resulting CompassJudger-2 series achieves superior performance on judge benchmarks, with our 7B model demonstrating competitive accuracy against significantly larger models like DeepSeek-V3-0324 \cite{liu2024deepseek} and Qwen3-235B-A22B \cite{qwen3}.


To advance the evaluation of judge models, we also present JudgerBenchV2, a standardized benchmark comprising 10,000 questions across 10 scenarios to evaluate judging capabilities. For the first time, it establishes category-specific judging standards and uses Mix-of-Judgers (MoJ) consensus as ground truth, paired with novel metrics that assess both sample-level accuracy and model-level rank consistency, providing a more robust evaluation.


To summarize, our contributions are as follows:
\setlist{nolistsep}
\begin{itemize}[leftmargin=1.5em]
    \item We develop a versatile, multi-styled judge data composition scheme with data curation and synthesis, enhancing CompassJudger-2’s robustness and domain adaptability at the data level.
    \item We significantly improve judge performance of \name{} by generating high-quality chain-of-thought judge data, selecting optimal training trajectories via rejection sampling, and applying policy gradient loss.
    \item We introduce JudgerBenchV2, which treats a Mix-of-Judgers  as ground truth and deploys new metrics that jointly assess accuracy and rank fidelity, enabling more reliable evaluation.
\end{itemize}

\section{Related Works}
\label{sec:related_work}


\para{LLM Judgers as Generative Verifiers.}
LLM-as-judge represents a novel approach where LLMs are fine-tuned to evaluate and provide judgment on model responses, offering not only a reward but also an analysis of the reasoning behind the decision. 
Unlike traditional reward models that assign a single reward value, LLMs can deliver more valuable feedback by explaining the logic and rationale of their judgments. 
However, many existing judge models~\cite{zhu2023judgelm, li2023generative} are trained for specific prompts, show poor generalization and cannot adapt to the diverse model evaluation needs. Therefore, all-in-one generative models have emerged, with CompassJudger-1~\cite{cao2024compassjudger} being the first to incorporate a wide range of judge tasks into model training, greatly enhancing the generalization ability. 
Con-J~\cite{ye2024scalarrewardmodellearning} and RISE~\cite{yu2025improve} have also conducted all-in-one Judge model training and achieved better Judge performance through the DPO strategy. 
Although these models have greatly ensured the generalization of prompts, they have not yet verified on other judge tasks such as critique generation and stylized judge.

\para{LLM Judging Evaluation.}Despite the rapid evolution of judge models, there is a notable lack of benchmarks for their evaluation. Rewardbench~\cite{lambert2024rewardbench} focuses on assessing a model’s reward capability across four categories: Chat, Chat Hard, Reasoning, and Safety. However, it faces issues with outdated data and a limited number of evaluation scenarios, leading to overfitting in many models on Rewardbench. JudgeBench~\cite{tan2024judgebench}, by contrast, evaluates judge models based on their ability to determine the correctness of answers in datasets like MMLU-Pro~\cite{wang2024mmlu} and LiveCodeBench~\cite{jain2024livecodebench}, thus testing their knowledge base to answer factual questions. RMB~\cite{zhou2024rmb} introduces a method using the Best of N (BoN) and involves a comparative model making multiple judgments to assess the consistency of the model’s judging. Nonetheless, these benchmarks only offer a limited view of the judging ability and do not encompass a wide enough range of evaluation scenarios.

\section{Methodology}
In this section, we first outline the training data pipeline for CompassJudger-2, covering data curation and data synthesis. We then explain how to apply rejection sampling and policy gradient optimization to incorporate verified rewards into the judge task. We apply the training data and the training strategy to the Qwen2.5-Instruct series of models, yielding our CompassJudger-2.

\begin{figure*}[!t]
    \centering
    \begin{minipage}[t]{1\textwidth}
        \centering
        \includegraphics[width=\textwidth]{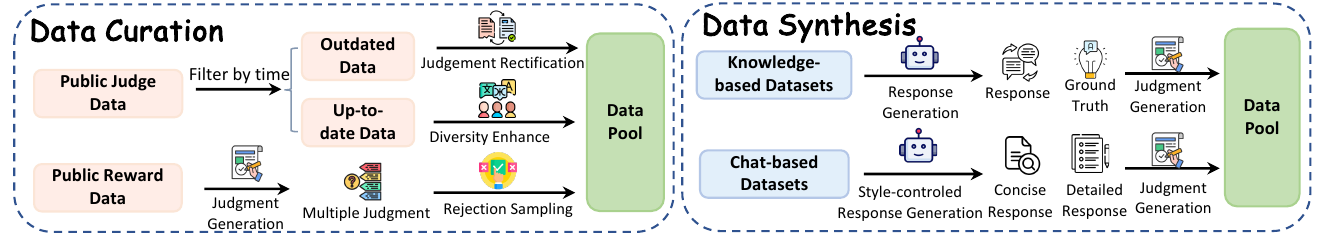}
        \caption{
            \small \textbf{The data construction pipeline of \name{}, including data curation and synthesis.} The Data Curation stage include reconstruction of public judge and reward data, while the data synthesis stage contains response generation over knowledge-based and chat based datasets.
        }
        \label{fig:data}
    \end{minipage}
    \hfill
\end{figure*}

\subsection{Overall Data Pipeline}
\para{Data Curation.}
We begin by collecting open-source judge-related datasets, including Public Judge Data and Public Reward Data. Public Judge Data  contain critiques and explanations while Public Reward Data only contain ground truth labels.

For \textit{\textbf{Public Judge Data}}, we observe that many judgments were generated by outdated models such as ChatGPT, which may introduce misjudgments and implicit errors. To address this issue, we split the data into \textit{outdated} and \textit{up-to-date} subsets based on the cutoff date of October 2024.
For \textit{{outdated data}}, we use Qwen2.5-72B-Instruct to reconstruct outdated judgment and further verify correctness by comparing the predictions with human-labeled ground truth, ensuring that only accurate judgments are preserved. For \textit{{up-to-date}} data, we leverage a large number of subjective evaluation datasets available in the community, such as ArenaHard \cite{li2024crowdsourced}, WildBench \cite{lin2024wildbench}, MTBench \cite{bai2024mt}, etc., to collect their judgment prompt templates, which are then used to replace the original prompt templates in the existing judgment data, thereby enhancing their diversity.

For \textit{\textbf{Public Reward Data}}, such data lacks critique annotations, making it suboptimal for training generative judge models. To leverage these data effectively, we prompt Qwen2.5-72B-Instruct to generate multiple judgments for each data instance and further refine the quality through rejection sampling. A detailed description of our construction pipeline is provided in Section \ref{sec:rft}.


\para{Data Synthesis.} 
To enhance the robustness and versatility, we systematically design and synthesize data from Knowledge-based Datasets and Chat-based Datasets, aiming to enrich world knowledge and improve stylistic adaptability, respectively. 

For \textit{\textbf{Knowledge-based Datasets}}, we aggregate model outputs from standardized benchmarks (e.g., MMLU\cite{wang2024mmlu}, CMMLU \cite{li2023cmmlu}, GSM8K \cite{cobbe2021training}) and employ Qwen2.5-72B-Instruct to evaluate their correctness while providing detailed rationales. These judgments are subsequently validated against ground truth answers, with only verified correct evaluations retained in the training corpus. 

For \textit{\textbf{Chat-based Datasets}}, we generate response pairs exhibiting contrasting characteristics  and instruct Qwen2.5-72B to select the superior response according to specified style requirements, thereby creating style-sensitive judgment data. 

\para{Overall Training Data Construction.} 
Prior studies~\cite{cao2024compassjudger, liu2025inference} have also demonstrated that incorporating general instruction data helps maintain a model's generalization capability while preserving its judge performance. Therefore, we also include general instruction data collected from CompassJudger-1 in our training dataset. The final training data for CompassJudger-2 consists of four components: (1) publicly available judge data that undergo diversity enhancement and quality rectification, (2) publicly available reward data process through rejection sampling (RFT data), (3) synthetic data generated from knowledge-based and chat-based datasets, and (4) general instruction data (G-SFT data). 

\begin{figure*}[!t]
    \centering
    \begin{minipage}[t]{1\textwidth}
        \centering
        \includegraphics[width=0.9\textwidth]{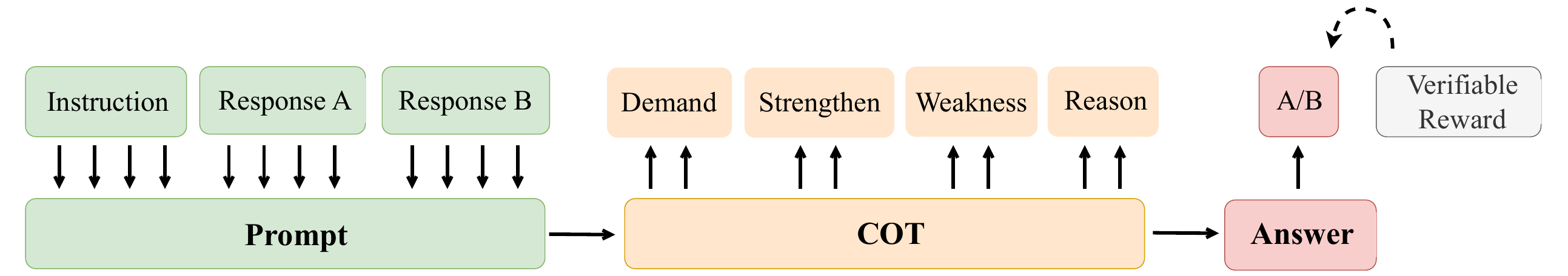}
        \caption{
            \small \textbf{Illustration of the reasoning path in the judge task.} The reasoning path involves critical analysis of the instruction and responses from various models. The final answer prediction can be treated as a classification task, which is further guided by a verified reward for supervision.
        }
        \label{fig:motivation}
    \end{minipage}
    \hfill
\end{figure*}
\begin{figure*}[!t]
    \centering    \includegraphics[width=0.9\textwidth]{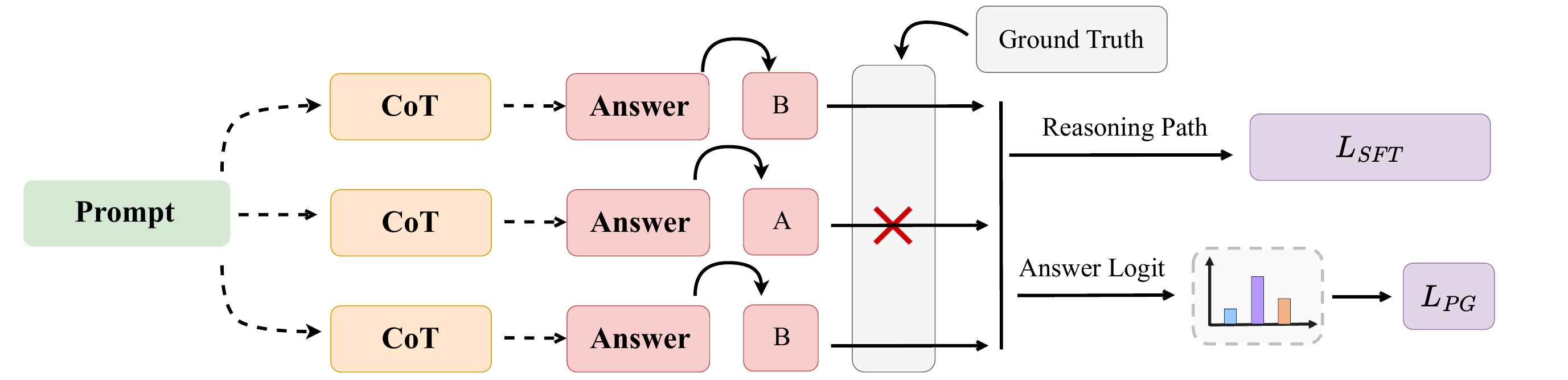}
    \vspace{-4pt}
    \caption{
        \small \textbf{Training framework of CompassJudger-2.} \name{} utilize rejection sampling to choose correct reasoning paths for SFT training and apply policy gradient loss over the answer logit to incorporate verifiable reward. 
    }
    \label{fig:eq}
\end{figure*}
\subsection{Incorporating Verified Reward}
\label{sec:rft}
To enhance the judge model's accuracy and generalization, we propose a training paradigm that integrates verifiable rewards through policy gradient optimization and rejection sampling. Specifically, we first guide the model to generate judgments through critical thinking, then reinforce the reward for final judgment outcomes using policy gradient loss, and further enhance judgment performance by incorporating a rejection sampling strategy. We elaborate on these steps in detail below.

\para{Critical Thinking.}
The SFT training of judge models requires high-quality instruction-response data, which can be costly to obtain. To tackle this challenge, we introduce an innovative chain-of-thought methodology aimed at producing high-quality instruction-response data specifically for the judge task. Following the reasoning pipeline in DeepSeek-R1 \cite{guo2025deepseek}, we craft a critical thinking prompt specifically for judge models, as shown in Figure~\ref{fig:motivation}. We divide the judge task into several important steps and require the model in making predictions through comprehensive thinking. Formally, the model is required to dissect the problem by evaluating: 
(1) User's Demand: The model need to analysis the specific requirements of the user's instruction.  (2) 
 Strengths of Model A/B. (3) Weaknesses of Model A/B. (4) Reasoning: Perform reasoning based on the aforementioned analysis.
(5) Prediction: Output the final prediction. 
In practice, we employ Qwen2.5-72B-Instruct \cite{yang2024qwen2} as backbone for data synthesis. 


\para{Judge Reward.}
In the judge task, the model performs binary classification by outputting its prediction at designated positions. This structured output enables us to utilize the ground truth labels as explicit guidance signals for optimization. Inspired by DeepSeek-R1 \cite{guo2025deepseek}, given an instruction-response pair $(x, y)$, a prediction position $k_x$, and the corresponding ground truth label $y_{k_x}^*$, we apply a rule-based reward $r(x, y)$ defined as 1 if the model’s prediction at position $k_x$ matches the ground truth label $y_{k_x}^*$, and 0 otherwise.

\para{Policy Gradient Optimization.}
We formulate the learning objective as maximizing the expected reward over the response distribution and 
the gradient of this objective can be derived as follows:

\begin{equation}
\begin{aligned}
& \nabla_{\theta} J(\theta) =\nabla_{\theta} \left[\mathbb{E}_{x \sim D} \mathbb{E}_{y \sim \pi_{\theta}} [r(x,y)]\right]\\
&= \mathbb{E}_{x \sim D} \mathbb{E}_{y \sim \pi_{\theta}} \left[ r(x,y) \nabla_{\theta} \log \pi_{\theta}(y|x) \right] \\
&= \mathbb{E}_{x \sim D} \mathbb{E}_{y \sim \pi_{\theta}} \left[ r(x,y) \sum_{t=1}^{n} \nabla_{\theta} \log \pi_{\theta}(y_t | x, y_{<t}) \right]
    \end{aligned}
\end{equation}
This decomposition shows how the gradient propagates through all sequence positions in autoregressive models.
Given that reward function only depends on the prediction at position $k_x$, the  policy gradient loss can be further simplified as follows:  
\begin{equation}
\begin{aligned}
\mathcal{L}_{\mathrm{PG}} 
= -\mathop{\mathbb{E}}\limits_{x\sim D} \mathop{\mathbb{E}}\limits_{y \sim \pi_{\theta}}\left[ \log \pi_{\theta}(y_{k_x} | x, y_{< {k_x}} ) \Big | _{y_{k_x} = y_{k_x}^*} \right].
\end{aligned}
\label{eq:pg_loss}
\end{equation}

We observe that SFT loss computes the conditional probability under fixed prefixes, while the policy gradient loss approximates the marginal probability by aggregating over diverse prefix.  
This distinction arises because SFT employs teacher forcing with deterministic prefixes, whereas policy gradient optimization explores various response trajectories to maximize expected rewards.

\para{Rejection Sampling for RL Generalization.}
While policy gradient optimization directly maximizes expected rewards, it suffers from limited exploration during the standard SFT stage, where fixed prefixes constrain the diversity of generated responses. To address this exploration bottleneck, we leverage rejection sampling to enhance model generalization through diversified prefix generation. 
Our approach systematically generates and filters diverse response candidates based on quality metrics and reject the samples that do not match the ground truth label. 
Formally, for the $i_{th} $  instruction $(i \in \{1..,N\})$ in the dataset,  we generate $M$ response samples that satisfy the ground truth label $y^{(i,*)}$ to approximate the policy gradient loss:
\begin{equation}
\begin{aligned}
L_{\mathrm{PG}} 
&= -\frac{1}{NM} \sum_{i=1}^{N} \sum_{j=1}^{M} \log \pi_{\theta} ( y_{k_x}^{(i,*)} | x^{(i)}, y_{< k_x}^{(i,*)} )
\end{aligned}
\end{equation}
Similarly, we apply the SFT loss to the sampled response candidates. We further combine the SFT loss and policy gradient loss:
\begin{equation}
\begin{aligned}
\mathcal{L}_{total} =& \mathcal{L}_{\mathrm{SFT}} + \mathcal{L}_{\mathrm{PG}}
\end{aligned}.
\end{equation}

\begin{table*}[!t]
    \centering
    \caption{
    \textbf{Total loss and mapping functions.} We discuss three mapping functions to approximate $g$ in the total loss.}\label{tab:mapping_loss}
    \begin{adjustbox}{width=\textwidth}
    \begin{tabular}{lc}
    \toprule
    \textbf{Loss} & \textbf{Loss Function} \\     
    \midrule
    Total Loss & $\mathcal{L}_{total} =-\frac{1}{NM} \sum_{i=1}^{N}\sum_{j=1}^{M}  [ \sum_{t \neq k_{ij}} \log \pi_{\theta} ( y_{t}^{(i,j)} | x^{(j)}, y_{< t}^{(i,j)} ) + g\left(\log \pi_{\theta} ( y_{k_{ij}}^{(i,*)} | x^{(j)}, y_{< k_{ij}}^{(i,j)} )\right) ]$\\
    
    \midrule
    \multicolumn{1}{l}{DPO Loss} & $
    \mathcal{L}_{\mathrm{DPO}} = 
    -\frac{1}{NM} \sum_{i=1}^{N}\sum_{j=1}^{M} \log \sigma\left(\beta \log \frac{\pi_{\theta}(y_{k_{ij}}^{(i,*)} | x^{(j)}, y_{< k_{ij}}^{(i,j)})}{\pi_{\theta}(y_{k_{ij}}^{(i,-)} | x^{(j)}, y_{< k_{ij}}^{(i,j)})}\right)$ \\
    \multicolumn{1}{l}{Temperature Loss} & $
    \mathcal{L}_{\mathrm{Temp}} = 
    -\frac{1}{NM} \sum_{i=1}^{N}\sum_{j=1}^{M} \log \frac{\exp(\log \pi_{\theta}(y_{k_{ij}}^{(i,*)} | x^{(j)}, y_{< k_{ij}}^{(i,j)}) / \tau)}{\sum_{y'} \exp(\log \pi_{\theta}(y' | x^{(j)}, y_{< k_{ij}}^{(i,j)}) / \tau)}
$ \\
    \multicolumn{1}{l}{Margin Loss}  & $
    \mathcal{L}_{\mathrm{Margin}} = 
    \frac{1}{NM} \sum_{i=1}^{N}\sum_{j=1}^{M} \max\big(0,  \gamma - \log \pi_{\theta}(y_{k_{ij}}^{(i,*)} | x^{(j)}, y_{< k_{ij}}^{(i,j)}) + \log \pi_{\theta}(y_{k_{ij}}^{(i,-)} | x^{(j)}, y_{< k_{ij}}^{(i,j)})\big)$ \\
    \bottomrule
    \end{tabular}
    \end{adjustbox}
\end{table*}
\para{Mapping Function.}
The total loss can be decomposed with SFT loss over the prefix and a mapping function $g$ over the prediction position, as shown in Table \ref{tab:mapping_loss}. We also design three different mapping loss function on the prediction position as $g$ for optimization over the ground truth answer $y_{k_{ij}}^{(i,*)}$ and the wrong answer $y_{k_{ij}}^{(i,-)}$. 

    \noindent $\bullet$ \textit{\para{DPO Loss w/o Reference Model}} encourages the model to increase the probability of true answer while decreasing the probability of wrong answer.

    \noindent$\bullet$ \textit{\para{Temperature Loss}} performs temperature scaling to the logits before softmax, effectively sharpening the probability distribution around the ground truth token with $\tau$ as the temperature.

\noindent$\bullet$ \textit{\para{Margin Loss}} introduces a margin $\gamma$ between the ground truth token and other answer, ensuring that the ground truth probability is sufficiently higher.


\begin{table*}[!t]
    \centering
    \caption{\textbf{Main results on judge benchmarks.} \name{} achieves state-of-the-art performance on both 7B and 32B+ variants. }
    \label{tab:main_results}
    \resizebox{.9\textwidth}{!}{
    \begin{tabular}{lcccccclcccclc}
    \toprule
    \textbf{Model} & \textbf{JudgerBenchV2} & \textbf{JudgeBench} & \textbf{RMB} & \textbf{RewardBench} & \textbf{Average} \\                          \midrule
    \multicolumn{6}{c}{General Models} \\ \midrule
    \multicolumn{1}{l}{Qwen2.5-7B-Instruct} \cite{yang2024qwen2} & 57.14 & 23.23 & 69.03 & 79.69 & 57.27 \\
    \multicolumn{1}{l}{Llama3.1-8B-Instruct} \cite{grattafiori2024llama} & 57.64 & 33.23 & 66.01 & 73.64 & 57.63 \\
    \multicolumn{1}{l}{InternLM3-8B-Instruct} \cite{cai2024internlm2} & 57.71 & 24.19 & 72.02 & 80.62 & 58.64 \\
    \multicolumn{1}{l}{Qwen2.5-32B-Instruct} \cite{yang2024qwen2}&  62.97 & 59.84 & 74.99 & 85.61 & 70.85 \\
    \multicolumn{1}{l}{DeepSeek-V3-0324} \cite{liu2024deepseek} & 64.43 & 59.68 & 78.16 & 85.17 & 71.86 \\
    \multicolumn{1}{l}{Qwen3-235B-A22B} \cite{qwen3}& 61.40 & 65.97 & 75.59 & 84.68 & 71.91 \\
    \midrule
\multicolumn{6}{c}{Reward Models} \\ 
\midrule
    \multicolumn{1}{l}{InternLM2-20B-reward} \cite{cai2024internlm2} & - & - &  62.90 & 90.20 & - \\
    \multicolumn{1}{l}{Deepseek-GRM-27B} \cite{liu2025inference} & - & - &  69.00 & 86.00 & - \\
    \multicolumn{1}{l}{RM-R1-Qwen-Instruct-32B} \cite{chen2025rm} & - & - &  73.00 & 92.90 & - \\
\midrule
    \multicolumn{6}{c}{7B Judge Models} \\ \midrule
    \multicolumn{1}{l}{CompassJudger-1-7B-Instruct} \cite{cao2024compassjudger} & 57.96 & 46.00 & 38.18 & 80.74 & 55.72 \\
    \multicolumn{1}{l}{Con-J-7B-Instruct} \cite{ye2025learning} & 52.35
 & 38.06
 & 71.50
 & 87.10
 & 62.25 \\
    \multicolumn{1}{l}{RISE-Judge-Qwen2.5-7B} \cite{yu2025improve} & 46.12 & 40.48 & 72.64 & 88.20 & 61.61 \\
    \rowcolor{tabhighlight}\multicolumn{1}{l}{\model{7B}} & \textbf{60.52} & \textbf{63.06} & \textbf{73.90} & \textbf{90.96} & \textbf{72.11} \\
    \midrule
    
    \multicolumn{6}{c}{32B+ Jugde Models} \\ \midrule
    \multicolumn{1}{l}{CompassJudger-1-32B-Instruct} \cite{cao2024compassjudger} & 60.33 & 62.29 & \textbf{77.63} & 86.17 & 71.61 \\
\multicolumn{1}{l}{Skywork-Critic-Llama-3.1-70B} \cite{skyworkcritic2024} & 52.41 & 50.65 & 65.50 & \textbf{93.30} & 65.47 \\
    
    \multicolumn{1}{l}{RISE-Judge-Qwen2.5-32B} \cite{yu2025improve} & 56.42 & 63.87 & 73.70 & 92.70 & 71.67 \\
    \rowcolor{tabhighlight}\multicolumn{1}{l}{\model{32B}} & \textbf{62.21} & \textbf{65.48} & 72.98 & 92.62 & \textbf{73.32} \\
    \bottomrule
        \end{tabular}
    }
\end{table*}

\begin{table*}[!t]
    \centering
    \small
    \caption{\textbf{Results on general benchmarks.} \name{} maintains strong performance on both objective and subjective datasets.}
    \label{tab:obj_sub}
    \resizebox{.9\textwidth}{!}{
    \begin{tabular}{lcccccclccccclc}
    \toprule
    \textbf{Model} & \textbf{MMLU Pro} & \textbf{GPQA Diamond} & \textbf{AIME2025} & \textbf{LiveCodeBench v5} & \textbf{IFEval} & \textbf{ArenaHard} \\                          \midrule
    \multicolumn{7}{c}{7B Judge Models} \\ \midrule
    \multicolumn{1}{l}{Qwen2.5-7B-Instruct} \cite{yang2024qwen2} & \textbf{55.43} & 34.85 & \textbf{6.67} & 12.57 & 73.20 & 47.86 \\
    \multicolumn{1}{l}{Con-J-7B-Instruct} \cite{ye2025learning}  & 44.74 &27.27  & 3.33 &  6.59 &54.90 & 23.49 \\
    \multicolumn{1}{l}{RISE-Judge-Qwen2.5-7B} \cite{yu2025improve} & 51.56 & 32.32 & \textbf{6.67} & 12.57 & 44.18 & 35.99 \\

    \rowcolor{tabhighlight}\multicolumn{1}{l}{\model{7B}} & {52.55} & \textbf{39.39} & \textbf{6.67} & \textbf{14.37} & \textbf{74.49} & \textbf{53.49} \\
    \midrule
    \multicolumn{7}{c}{32B Judge Models} \\ \midrule
    \multicolumn{1}{l}{Qwen2.5-32B-Instruct} \cite{yang2024qwen2} & 68.92 & 42.93 & \textbf{16.67} & \textbf{30.54} & 79.85 & 70.16 \\
    \multicolumn{1}{l}{RISE-Judge-Qwen2.5-32B} \cite{yu2025improve} & 67.88 & 42.93 & 6.67 & 27.54 & 62.85 & 61.52 \\
    \rowcolor{tabhighlight}\multicolumn{1}{l}{CompassJudger-2-32B-Instruct} & \textbf{69.22} & \textbf{50.51} & \textbf{16.67} & 25.15 & \textbf{79.48} & \textbf{83.31} \\
    \bottomrule
        \end{tabular}
    }
    \label{tab:type}
\end{table*}

\section{JudgerBenchV2: A More Robust Benchmark for Judge Models}
Existing benchmarks for judge models have numerous limitations, such as insufficient coverage of judge scenarios and a lack of sufficiently accurate ground truth (GT). 
To address these issues, we propose JudgerBenchV2, aiming to improve the evaluation landscape for judge models and provide a more comprehensive and accurate benchmark.

\para{Data Construction.}
We first collect real-world user queries in Chinese and English through CompassArena \cite{2023opencompass}, and cluster them via K-means. We then utilize an LLM to classifies each query by difficulty level and manually select 100 queries per scenario, ensuring a balanced distribution of languages and difficulty level.
Next, we select 10 high-performing models of comparable capability and generate their responses to these queries. 
We then use GPT-4o-mini as the policy model and pair it with each of the 10 models to form response pairs. A judge model evaluates these pairs in a pairwise manner to obtain judge results. By comparing with the GT, we derive the performance scores of the judge model.

\para{Mixture of Judges.}
Evaluating open-ended questions is highly subjective since different individuals may produce varying judgments, and different models also exhibit judge biases. Relying solely on the judgments from a single human or a single model as GT thus risks introducing bias.
To address this, we introduce the mixture of judgers (MoJ) strategy, leveraging the judgments of DeepSeek-R1, DeepSeek-v3-0324, and Qwen3-235B-A22B and their majority consensus is considered as GT.

\para{Robust Judge Performance Metrics.}
Traditional judge evaluation metrics primarily focus on sample-level accuracy and fail to capture essential dimensions like ranking consistency. For example, human raters often converge on overall model rankings although they may disagree on individual samples. 
A comprehensive evaluation framework should therefore incorporate both fine-grained judgment accuracy and high-level ranking fidelity.

In JudgerBenchV2, we conduct pairwise comparisons between a candidate model and GPT-4o-mini to determine which delivers superior responses. Each comparison is evaluated by both a ground truth judge model and a test judge model. A sample is considered correct if both judges agree on the better-performing model. For each sample, the model deemed superior earns a score increment of 1. The total number of pairwise samples is denoted by \( N \) and \( C \) represents the number of samples where the GT and test judge models agree on the superior model.
For a set of \( M \) candidate models, let the GT judge model and the test judge model generate score lists \( S_1 = \{s_{1,m}\}_{m \in M} \) and \( S_2 = \{s_{2,m}\}_{m \in M} \), respectively, where \( s_{i,m} \) represents the cumulative score for model \( m \) based on pairwise wins. Additionally, let \( R_1 = \{r_{1,m}\}_{m \in M} \) and \( R_2 = \{r_{2,m}\}_{m \in M} \) denote the rank lists, where \( r_{i,m} \) is the rank of model \( m \) according to judge \( i \). The performance of the test judge model is evaluated using the following metric:

\begin{equation}
\begin{aligned}
\mathcal{P} = 100 \cdot \underbrace{ \frac{C}{N}}_{ \substack{\text{Sample-level} \\ \text{accuracy}}} 
& - \frac{100}{|M|}  \sum_{m \in M} ( *\underbrace{\frac{|r_{1,m} - r_{2,m}|}{|M| - 1}}_{\substack{\text{Normalized rank} \\ \text{difference}}} 
& + \underbrace{\frac{|s_{1,m} - s_{2,m}|}{\max_{m' \in M} |s_{1,m'} - s_{2,m'}|}}_{\substack{\text{Normalized score difference}}} ).
\end{aligned}
\end{equation}
The first term  captures the sample-level accuracy by measuring the agreements between the judges. The second term penalizes discrepancies in rankings and scores, 
with normalization to ensure equitable comparisons across different models.


\begin{figure*}[!t]
	\centering
	\begin{subfigure}{0.45\linewidth}
		\includegraphics[width=\linewidth]{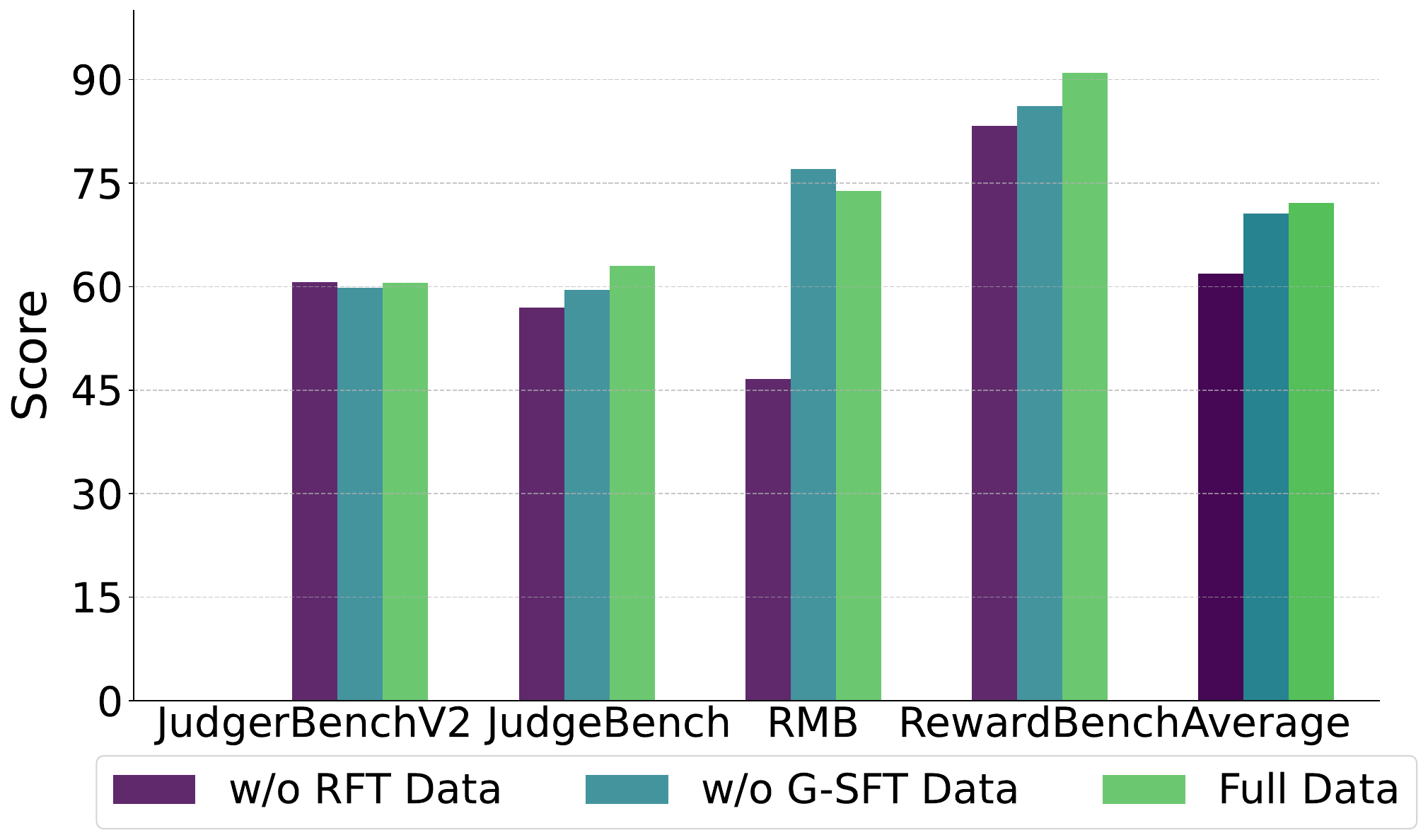}
		\caption{Judge benchmarks. }
	\end{subfigure}
    \hfill
	\begin{subfigure}{0.45\linewidth}
		\includegraphics[width=\linewidth]{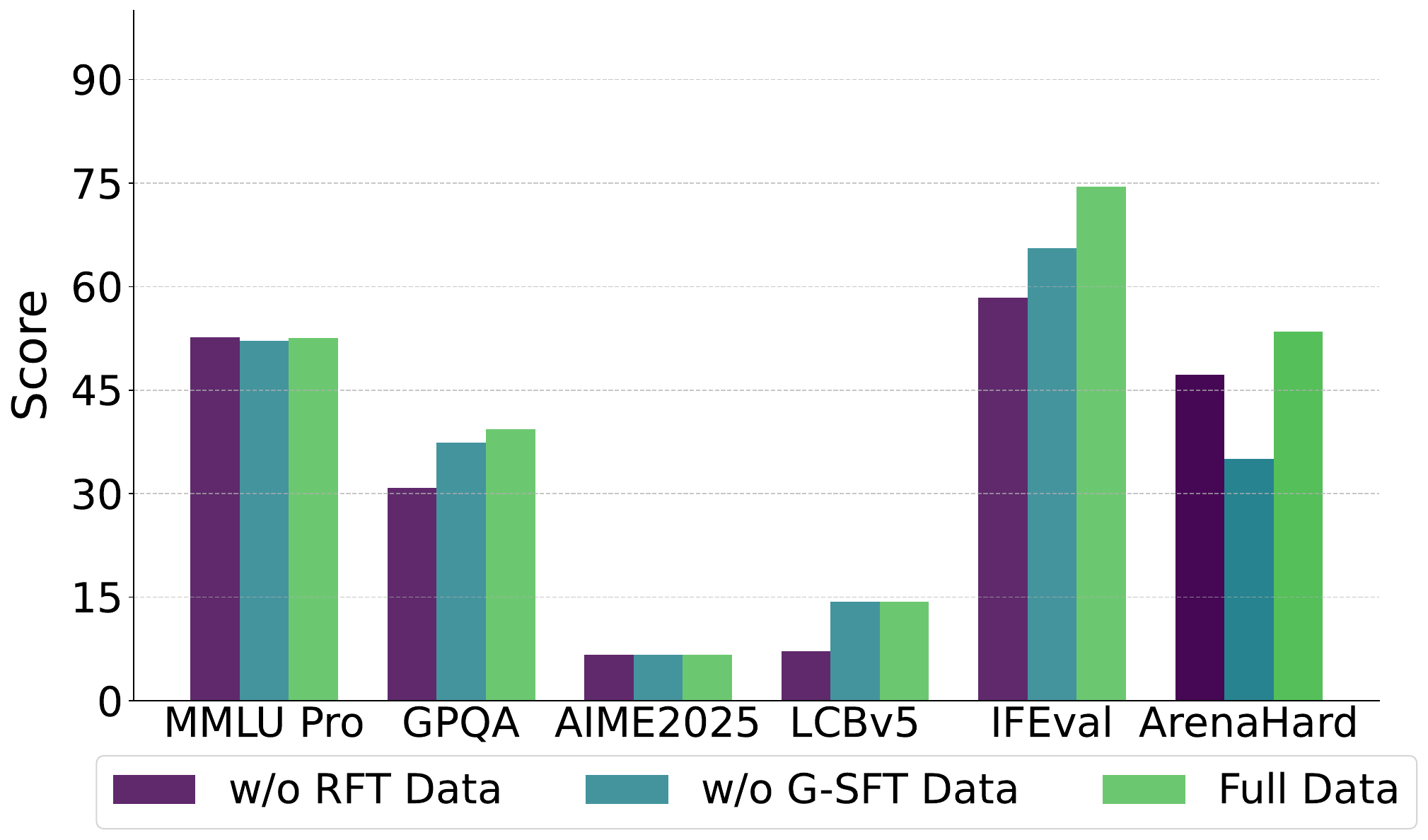}
		\caption{Subjective and objective benchmarks.}
	\end{subfigure}
    \hfill
	\caption{\textbf{Data ablation results on different benchmarks.} 
    }
	\label{fig:data_ablation}
\end{figure*}

\section{Experiments}
\label{sec:experiments}

\subsection{Experimental Setup}
\label{subsec:setup}
\para{Evaluation Datasets.}
We evaluate the performance of \name{} on leading judge benchmarks, including RewardBench \cite{lambert2024rewardbench}, JudgeBench \cite{tan2024judgebench}, RMB \cite{zhou2024rmb}, and our JudgerBenchV2. Besides, we compare our method with other judge models over popular objective and subjective benchmarks, including MMLU Pro \cite{wang2024mmlu}, GPQA Diamond \cite{rein2024gpqa}, AIME2025, LiveCodeBench v5 \cite{jain2024livecodebench}, IFEval \cite{ding2023enhancing} and ArenaHard \cite{li2024crowdsourced}. We further conduct extensive experiments on AlignBench \cite{liu2023alignbench} and AlpacaEval \cite{dubois2024length}, showcasing the critique ability of \name{} for model improvement.

\para{Training Settings.}
\label{sec:details}
In practice we generate 8 candidate responses for filtering during rejection sampling. 
For model training, we utilize Qwen-2.5 series as the checkpoint and adopt 6e-5 as the learning rate. For policy gradient loss parameter, we set $\beta=0.1$ in DPO loss, $\tau=5$ in temperature loss and $\gamma=10$ in margin loss. We apply DPO loss on only the candidate answer and margin loss on the top 10 logits. We train the model for 1 epoch with batch size equal to 512.

\begin{table}[!t]
    \centering
    \caption{\textbf{Ablation results with policy gradient loss on CompassJudger-2-7B-Instruct.} Margin loss provides a significant boost compared to other forms of loss.}\label{tab:loss}
    \begin{adjustbox}{width=0.9\textwidth}
    \begin{tabular}{lcccccclcccclc}
    \toprule
    \textbf{Loss} & \textbf{JudgerBenchV2} & \textbf{JudgeBench} & \textbf{RMB} & \textbf{RewardBench} & \textbf{Average} \\                          \midrule
    \multicolumn{1}{l}{Baseline} & 60.20 & 61.77 & 68.13 & 89.50 & 69.90 \\
    \multicolumn{1}{l}{DPO} & 60.56 & 61.13 & 66.35 & 90.07 & 69.53 \\
    \multicolumn{1}{l}{Temperature}  & 59.43 & 62.42 & 67.77 & 90.25 & 69.97 \\
    \rowcolor{tabhighlight}\multicolumn{1}{l}{Margin} & 60.52 & 63.06 & 73.90 & 90.96 & 72.11 \\
    \bottomrule
    \end{tabular}
    \end{adjustbox}
\end{table}

\subsection{Main Results}

\para{Judge Ability Analysis.}
To verify the judge ability of our method, we conduct evaluation across multiple benchmarks and compare our method with general models, reward models and specialized judge models including the Skywork \cite{skyworkcritic2024} and RISE \cite{yu2025improve} series. As presented in Table~\ref{tab:main_results}, CompassJudger-2 consistently surpasses all baselines in average performance, demonstrating significant advancements in the generalization ability. Notably, CompassJudger-2-7B-Instruct outperforms RISE-Judge-Qwen2.5-7B by 22.58\% on JudgeBench and by 10.5\% on average. Compared to the CompassJudger-1 series, CompassJudger-2 enhances judge performance by 16.39\% for the 7B model and 1.71\% for the 32B model, on average.

\para{General Ability Analysis.}
We further highlight the improvements in general capabilities of \name{} compared to other judge models across objective and subjective benchmarks, as shown in Table~\ref{tab:obj_sub}. \name{} achieves markedly superior performance over other judge models on both objective and subjective datasets, demonstrating its generalization ability. Remarkably, \name{} surpasses general models like Qwen2.5-7B-Instruct and Qwen2.5-32B-Instruct on specific datasets, revealing a strong correlation between judge ability and general ability in LLMs and their potential to enhance each other.

\begin{table}[!t]
    \centering
    \caption{\textbf{Model improvement with generated critique on chat-based datasets.} AlignBench scores range from 0 to 10 and other datasets score range from 0-100. To standardize the scale, we normalize all the scores to a 0–100 range and then compute the average. }
    \label{tab:critique}
    \resizebox{.8\textwidth}{!}{
    \begin{tabular}{lcccc}
    \toprule
    \textbf{Model} & \textbf{AlignBench} & \textbf{AlpacaEval} & \textbf{ArenaHard} & \textbf{Average} \\                          \midrule
    \multicolumn{5}{c}{Policy Model: LLama3.1-8B-Instruct} \\ \midrule
    \multicolumn{1}{l}{Base}  & 4.90 & 27.95 & 29.11 & 35.35 \\
    \multicolumn{1}{l}{RISE-Judge-Qwen2.5-7B} & 4.99 & 28.03 & 28.64 & 35.52  \\
    \rowcolor{tabhighlight}\multicolumn{1}{l}{\model{7B}} & 5.20 & 30.68 & 32.76 & 38.48 \\
    \midrule
    \multicolumn{5}{c}{Policy Model: Qwen2.5-7B-Instruct} \\ \midrule
    \multicolumn{1}{l}{Base} & 6.65 & 36.65 & 47.86 & 50.34 \\
    \multicolumn{1}{l}{RISE-Judge-Qwen2.5-7B} & 6.43&35.12&45.07&48.16  \\
    \rowcolor{tabhighlight}\multicolumn{1}{l}{\model{7B}} & 6.76 & 38.14 & 51.15 & 52.30 \\
    \midrule
    \multicolumn{5}{c}{Policy Model: InternLM3-8B-Instruct} \\ \midrule
    \multicolumn{1}{l}{Base} & 6.46 & 64.84 & 46.27 & 58.57 \\
    \multicolumn{1}{l}{RISE-Judge-Qwen2.5-7B} &6.47&62.17&43.89&56.92  \\
    \rowcolor{tabhighlight}\multicolumn{1}{l}{\model{7B}} & 6.50 & 65.85 & 47.76 & 59.54 \\
    \bottomrule
        \end{tabular}
    }
\end{table}

\begin{figure*}[!t]
    \centering    \includegraphics[width=.9\textwidth]{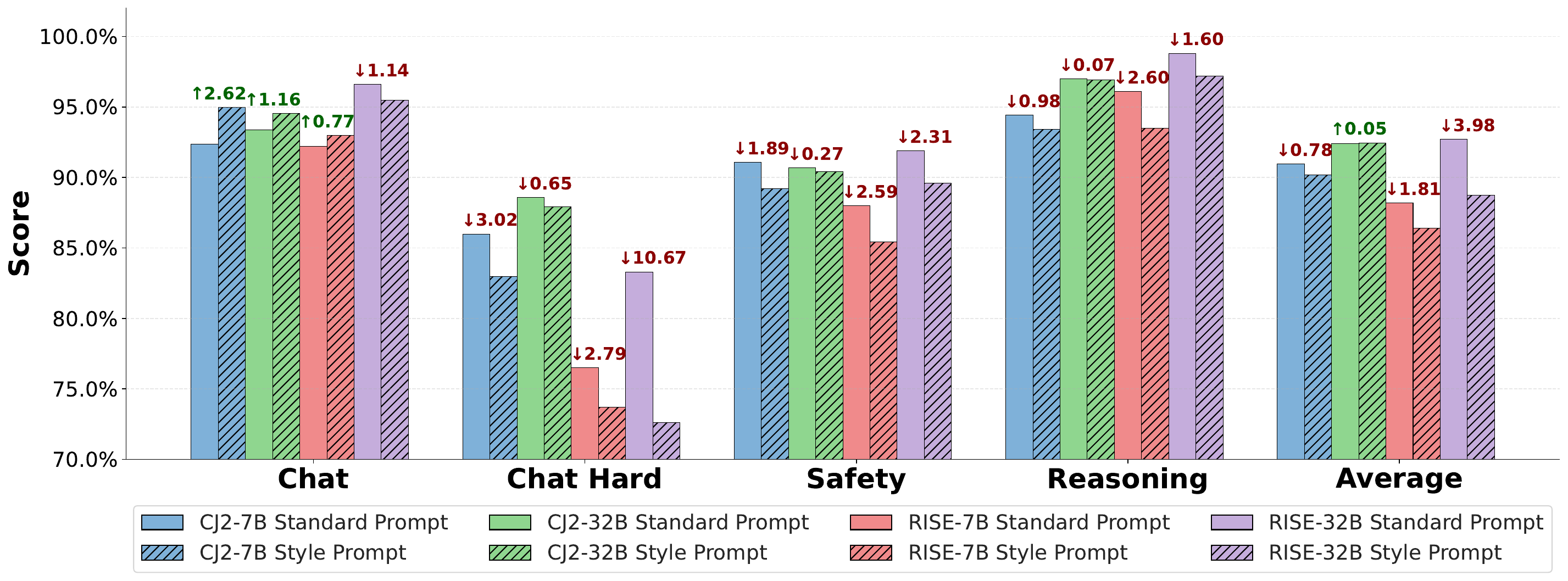}
    \caption{\textbf{
        Comparison results over style judge of \name{} and RISE.}
    }
    \label{fig:style}
\end{figure*}

\subsection{Ablation Study}

\para{Policy Gradient Loss.}
To evaluate the impact of incorporating policy gradient loss, we conduct a thorough ablation study to determine the most effective type of policy gradient loss for improving model performance. We compare a baseline model without policy gradient loss to models using various policy gradient losses, including DPO, Temperature, and Margin loss on 7B level, as presented in Table~\ref{tab:loss}. Our findings reveal that the verified reward serves as a critical supervised signal, significantly enhancing the model’s performance in judge tasks. All models with policy gradient loss surpass the baseline model on RewardBench, achieving performance improvements ranging from 0.5\% to 1.4\%. Notably, the model with margin loss demonstrate the best generalization across 3 out of 4 datasets, delivering an 2.21\% performance on average boost compared to the baseline model. As a result, we select margin loss as the default choice for our study.

\para{Data Ablation.}
To investigate how general instruction data (G-SFT Data) and rejection sampling (RFT Data) impact judge ability and general ability, we perform ablation studies by separately removing each data type from the training set. As illustrated in Figure~\ref{fig:data_ablation}, the results highlight several key findings. Removing RFT data causes a significant decline in judge performance, mainly due to lower judge consistency and result in poor results on the RMB dataset. In addition, including RFT data enhances performance across specific datasets, such as GPQA-Diamond and ArenaHard, underscoring its role in boosting general ability. In contrast, General SFT data primarily maintain the general ability of the model, with minimal impact on judge ability.


\subsection{Discussions}

\para{Critique Ability for Model Improvement.} An effective all-in-one judge model should be capable to produce high-quality critiques that offer insightful analysis and explanations. To evaluate the critique ability of \name, we task it with generating analyses of responses from various policy models on subjective datasets. We then permit the policy models to revise their initial responses based on these critiques. For comparison, we present the initial scores of the policy models (Base) alongside the results of using RISE-Judge-Qwen2.5-7B as the critique model, as shown in Table~\ref{tab:critique}. 
The results reveal a striking insight that all policy models improve when guided by critiques from \name{}, whereas low-quality critiques from RISE-Judge-Qwen2.5-7B often result in performance drop. This suggest the superior critique quality of \name{} and highlights its potential to enhance training performance during model iterations. 
We also provide some case study for comparison in the Appendix.

\para{Style Judge.} An effective all-in-one judge model should also maintain consistent performance with various prompts. 
Therefore, we conduct style judge experiment with modifying judging prompts by adding following sentences: "Beyond this, users prefer a more detailed response; therefore, you need to determine which model’s answer provides more comprehensive and useful information when both responses are correct and have completed the user’s request".
We present the results on different subset of RewardBench. 
As can be seen from the results in Figure~\ref{fig:style}, RISE-32B suffers from significant performance drop by 10.67\% in the Chat Hard subset. 
Compared with RISE, \name{} are less sensitive of judging prompts and show better consistency and generalization ability, indicating the superiority of our method. 

\section{Conclusions}
\label{sec:conslusions}
In this work, we present CompassJudger-2, an series of all-in-one judge models that advance LLM-as-judge performance through a unified training paradigm combining diverse task-driven data composition, high-quality chain-of-thought supervision, and verifiable reward-guided optimization. 
Furthermore, we introduce JudgerBenchV2, a comprehensive benchmark with mixed-of-judgers and novel ranking-aware metrics, to enable more nuanced and reliable evaluation of judge models. Looking forward, CompassJudger-2 paves the way for more adaptable, interpretable, and efficient judge services in real-world LLM deployments, and we anticipate that extending this work to multi-modal and interactive evaluation scenarios will further enhance its applicability and impact.


\section{Limitations}
Despite the superior performance, there are still some limitations of \name{}. Rejection sampling incurs relatively higher inference costs, and the hallucinations produced by the LLM when synthesizing data may pose potential risks. These issues need to be further addressed, which will in turn enhance the performance of the judge models.

\bibliographystyle{plainnat}
\bibliography{main}

\begin{thebibliography}{35}
\providecommand{\natexlab}[1]{#1}
\providecommand{\url}[1]{\texttt{#1}}
\expandafter\ifx\csname urlstyle\endcsname\relax
  \providecommand{\doi}[1]{doi: #1}\else
  \providecommand{\doi}{doi: \begingroup \urlstyle{rm}\Url}\fi

\bibitem[Bai et~al.(2024)Bai, Liu, Bu, He, Liu, Zhou, Lin, Su, Ge, Zheng, et~al.]{bai2024mt}
Ge~Bai, Jie Liu, Xingyuan Bu, Yancheng He, Jiaheng Liu, Zhanhui Zhou, Zhuoran Lin, Wenbo Su, Tiezheng Ge, Bo~Zheng, et~al.
\newblock Mt-bench-101: A fine-grained benchmark for evaluating large language models in multi-turn dialogues.
\newblock \emph{arXiv preprint arXiv:2402.14762}, 2024.

\bibitem[Cai et~al.(2024)Cai, Cao, Chen, Chen, Chen, Chen, Chen, Chen, Chen, Chu, Dong, Duan, Fan, Fei, Gao, Ge, Gu, Gu, Gui, Guo, Guo, He, Hu, Huang, Jiang, Jiao, Jin, Lei, Li, Li, Li, Li, Li, Li, Liu, Liu, Hong, Liu, Liu, Liu, Lv, Lv, Lv, Ma, Ma, Ma, Ning, Ouyang, Qiu, Qu, Shang, Shao, Song, Song, Sui, Sun, Sun, Tang, Wang, Wang, Wang, Wang, Wang, Wang, Wang, Wei, Weng, Wu, Xiong, Xu, Xu, Yan, Yan, Yang, Ye, Ying, Yu, Yu, Zang, Zhang, Zhang, Zhang, Zhang, Zhang, Zhang, Zhang, Zhang, Zhang, Zhang, Zhang, Zhao, Zhao, Zhao, Zhou, Zhou, Zhuo, Zou, Qiu, Qiao, and Lin]{cai2024internlm2}
Zheng Cai, Maosong Cao, Haojiong Chen, Kai Chen, Keyu Chen, Xin Chen, Xun Chen, Zehui Chen, Zhi Chen, Pei Chu, Xiaoyi Dong, Haodong Duan, Qi~Fan, Zhaoye Fei, Yang Gao, Jiaye Ge, Chenya Gu, Yuzhe Gu, Tao Gui, Aijia Guo, Qipeng Guo, Conghui He, Yingfan Hu, Ting Huang, Tao Jiang, Penglong Jiao, Zhenjiang Jin, Zhikai Lei, Jiaxing Li, Jingwen Li, Linyang Li, Shuaibin Li, Wei Li, Yining Li, Hongwei Liu, Jiangning Liu, Jiawei Hong, Kaiwen Liu, Kuikun Liu, Xiaoran Liu, Chengqi Lv, Haijun Lv, Kai Lv, Li~Ma, Runyuan Ma, Zerun Ma, Wenchang Ning, Linke Ouyang, Jiantao Qiu, Yuan Qu, Fukai Shang, Yunfan Shao, Demin Song, Zifan Song, Zhihao Sui, Peng Sun, Yu~Sun, Huanze Tang, Bin Wang, Guoteng Wang, Jiaqi Wang, Jiayu Wang, Rui Wang, Yudong Wang, Ziyi Wang, Xingjian Wei, Qizhen Weng, Fan Wu, Yingtong Xiong, Chao Xu, Ruiliang Xu, Hang Yan, Yirong Yan, Xiaogui Yang, Haochen Ye, Huaiyuan Ying, Jia Yu, Jing Yu, Yuhang Zang, Chuyu Zhang, Li~Zhang, Pan Zhang, Peng Zhang, Ruijie Zhang, Shuo Zhang, Songyang Zhang, Wenjian Zhang,
  Wenwei Zhang, Xingcheng Zhang, Xinyue Zhang, Hui Zhao, Qian Zhao, Xiaomeng Zhao, Fengzhe Zhou, Zaida Zhou, Jingming Zhuo, Yicheng Zou, Xipeng Qiu, Yu~Qiao, and Dahua Lin.
\newblock Internlm2 technical report, 2024.

\bibitem[Cao et~al.(2024)Cao, Lam, Duan, Liu, Zhang, and Chen]{cao2024compassjudger}
Maosong Cao, Alexander Lam, Haodong Duan, Hongwei Liu, Songyang Zhang, and Kai Chen.
\newblock Compassjudger-1: All-in-one judge model helps model evaluation and evolution.
\newblock \emph{arXiv preprint arXiv:2410.16256}, 2024.

\bibitem[Chen et~al.(2025{\natexlab{a}})Chen, Yu, Wang, Zhang, Tang, Xiong, Li, Yang, and Li]{chen2025xverify}
Ding Chen, Qingchen Yu, Pengyuan Wang, Wentao Zhang, Bo~Tang, Feiyu Xiong, Xinchi Li, Minchuan Yang, and Zhiyu Li.
\newblock xverify: Efficient answer verifier for reasoning model evaluations.
\newblock \emph{arXiv preprint arXiv:2504.10481}, 2025{\natexlab{a}}.

\bibitem[Chen et~al.(2025{\natexlab{b}})Chen, Li, Wang, Jin, Qian, Wang, Wang, Zhang, Zhang, Zhang, et~al.]{chen2025rm}
Xiusi Chen, Gaotang Li, Ziqi Wang, Bowen Jin, Cheng Qian, Yu~Wang, Hongru Wang, Yu~Zhang, Denghui Zhang, Tong Zhang, et~al.
\newblock Rm-r1: Reward modeling as reasoning.
\newblock \emph{arXiv preprint arXiv:2505.02387}, 2025{\natexlab{b}}.

\bibitem[Clark et~al.(2019)Clark, Lee, Chang, Kwiatkowski, Collins, and Toutanova]{clark2019boolqexploringsurprisingdifficulty}
Christopher Clark, Kenton Lee, Ming-Wei Chang, Tom Kwiatkowski, Michael Collins, and Kristina Toutanova.
\newblock Boolq: Exploring the surprising difficulty of natural yes/no questions, 2019.
\newblock URL \url{https://arxiv.org/abs/1905.10044}.

\bibitem[Cobbe et~al.(2021)Cobbe, Kosaraju, Bavarian, Chen, Jun, Kaiser, Plappert, Tworek, Hilton, Nakano, et~al.]{cobbe2021training}
Karl Cobbe, Vineet Kosaraju, Mohammad Bavarian, Mark Chen, Heewoo Jun, Lukasz Kaiser, Matthias Plappert, Jerry Tworek, Jacob Hilton, Reiichiro Nakano, et~al.
\newblock Training verifiers to solve math word problems.
\newblock \emph{arXiv preprint arXiv:2110.14168}, 2021.

\bibitem[Contributors(2023)]{2023opencompass}
OpenCompass Contributors.
\newblock Opencompass: A universal evaluation platform for foundation models.
\newblock \url{https://github.com/open-compass/opencompass}, 2023.

\bibitem[Ding et~al.(2023)Ding, Chen, Xu, Qin, Zheng, Hu, Liu, Sun, and Zhou]{ding2023enhancing}
Ning Ding, Yulin Chen, Bokai Xu, Yujia Qin, Zhi Zheng, Shengding Hu, Zhiyuan Liu, Maosong Sun, and Bowen Zhou.
\newblock Enhancing chat language models by scaling high-quality instructional conversations.
\newblock \emph{arXiv preprint arXiv:2305.14233}, 2023.

\bibitem[Dua et~al.(2019)Dua, Wang, Dasigi, Stanovsky, Singh, and Gardner]{dua2019dropreadingcomprehensionbenchmark}
Dheeru Dua, Yizhong Wang, Pradeep Dasigi, Gabriel Stanovsky, Sameer Singh, and Matt Gardner.
\newblock Drop: A reading comprehension benchmark requiring discrete reasoning over paragraphs, 2019.
\newblock URL \url{https://arxiv.org/abs/1903.00161}.

\bibitem[Dubois et~al.(2024)Dubois, Galambosi, Liang, and Hashimoto]{dubois2024length}
Yann Dubois, Bal{\'a}zs Galambosi, Percy Liang, and Tatsunori~B Hashimoto.
\newblock Length-controlled alpacaeval: A simple way to debias automatic evaluators.
\newblock \emph{arXiv preprint arXiv:2404.04475}, 2024.

\bibitem[Grattafiori et~al.(2024)Grattafiori, Dubey, Jauhri, Pandey, Kadian, Al-Dahle, Letman, Mathur, Schelten, Vaughan, et~al.]{grattafiori2024llama}
Aaron Grattafiori, Abhimanyu Dubey, Abhinav Jauhri, Abhinav Pandey, Abhishek Kadian, Ahmad Al-Dahle, Aiesha Letman, Akhil Mathur, Alan Schelten, Alex Vaughan, et~al.
\newblock The llama 3 herd of models.
\newblock \emph{arXiv preprint arXiv:2407.21783}, 2024.

\bibitem[Guo et~al.(2025)Guo, Yang, Zhang, Song, Zhang, Xu, Zhu, Ma, Wang, Bi, et~al.]{guo2025deepseek}
Daya Guo, Dejian Yang, Haowei Zhang, Junxiao Song, Ruoyu Zhang, Runxin Xu, Qihao Zhu, Shirong Ma, Peiyi Wang, Xiao Bi, et~al.
\newblock Deepseek-r1: Incentivizing reasoning capability in llms via reinforcement learning.
\newblock \emph{arXiv preprint arXiv:2501.12948}, 2025.

\bibitem[Jaech et~al.(2024)Jaech, Kalai, Lerer, Richardson, El-Kishky, Low, Helyar, Madry, Beutel, Carney, et~al.]{jaech2024openai}
Aaron Jaech, Adam Kalai, Adam Lerer, Adam Richardson, Ahmed El-Kishky, Aiden Low, Alec Helyar, Aleksander Madry, Alex Beutel, Alex Carney, et~al.
\newblock Openai o1 system card.
\newblock \emph{arXiv preprint arXiv:2412.16720}, 2024.

\bibitem[Jain et~al.(2024)Jain, Han, Gu, Li, Yan, Zhang, Wang, Solar-Lezama, Sen, and Stoica]{jain2024livecodebench}
Naman Jain, King Han, Alex Gu, Wen-Ding Li, Fanjia Yan, Tianjun Zhang, Sida Wang, Armando Solar-Lezama, Koushik Sen, and Ion Stoica.
\newblock Livecodebench: Holistic and contamination free evaluation of large language models for code.
\newblock \emph{arXiv preprint arXiv:2403.07974}, 2024.

\bibitem[Lambert et~al.(2024)Lambert, Pyatkin, Morrison, Miranda, Lin, Chandu, Dziri, Kumar, Zick, Choi, et~al.]{lambert2024rewardbench}
Nathan Lambert, Valentina Pyatkin, Jacob Morrison, LJ~Miranda, Bill~Yuchen Lin, Khyathi Chandu, Nouha Dziri, Sachin Kumar, Tom Zick, Yejin Choi, et~al.
\newblock Rewardbench: Evaluating reward models for language modeling.
\newblock \emph{arXiv preprint arXiv:2403.13787}, 2024.

\bibitem[Li et~al.(2023{\natexlab{a}})Li, Zhang, Koto, Yang, Zhao, Gong, Duan, and Baldwin]{li2023cmmlu}
Haonan Li, Yixuan Zhang, Fajri Koto, Yifei Yang, Hai Zhao, Yeyun Gong, Nan Duan, and Timothy Baldwin.
\newblock Cmmlu: Measuring massive multitask language understanding in chinese.
\newblock \emph{arXiv preprint arXiv:2306.09212}, 2023{\natexlab{a}}.

\bibitem[Li et~al.(2023{\natexlab{b}})Li, Sun, Yuan, Fan, Zhao, and Liu]{li2023generative}
Junlong Li, Shichao Sun, Weizhe Yuan, Run-Ze Fan, Hai Zhao, and Pengfei Liu.
\newblock Generative judge for evaluating alignment.
\newblock \emph{arXiv preprint arXiv:2310.05470}, 2023{\natexlab{b}}.

\bibitem[Li et~al.(2024)Li, Chiang, Frick, Dunlap, Wu, Zhu, Gonzalez, and Stoica]{li2024crowdsourced}
Tianle Li, Wei-Lin Chiang, Evan Frick, Lisa Dunlap, Tianhao Wu, Banghua Zhu, Joseph~E Gonzalez, and Ion Stoica.
\newblock From crowdsourced data to high-quality benchmarks: Arena-hard and benchbuilder pipeline.
\newblock \emph{arXiv preprint arXiv:2406.11939}, 2024.

\bibitem[Lin et~al.(2024)Lin, Deng, Chandu, Brahman, Ravichander, Pyatkin, Dziri, Bras, and Choi]{lin2024wildbench}
Bill~Yuchen Lin, Yuntian Deng, Khyathi Chandu, Faeze Brahman, Abhilasha Ravichander, Valentina Pyatkin, Nouha Dziri, Ronan~Le Bras, and Yejin Choi.
\newblock Wildbench: Benchmarking llms with challenging tasks from real users in the wild.
\newblock \emph{arXiv preprint arXiv:2406.04770}, 2024.

\bibitem[Liu et~al.(2024)Liu, Feng, Xue, Wang, Wu, Lu, Zhao, Deng, Zhang, Ruan, et~al.]{liu2024deepseek}
Aixin Liu, Bei Feng, Bing Xue, Bingxuan Wang, Bochao Wu, Chengda Lu, Chenggang Zhao, Chengqi Deng, Chenyu Zhang, Chong Ruan, et~al.
\newblock Deepseek-v3 technical report.
\newblock \emph{arXiv preprint arXiv:2412.19437}, 2024.

\bibitem[Liu et~al.(2023)Liu, Lei, Wang, Huang, Feng, Wen, Cheng, Ke, Xu, Tam, et~al.]{liu2023alignbench}
Xiao Liu, Xuanyu Lei, Shengyuan Wang, Yue Huang, Zhuoer Feng, Bosi Wen, Jiale Cheng, Pei Ke, Yifan Xu, Weng~Lam Tam, et~al.
\newblock Alignbench: Benchmarking chinese alignment of large language models.
\newblock \emph{arXiv preprint arXiv:2311.18743}, 2023.

\bibitem[Liu et~al.(2025)Liu, Wang, Xu, Ma, Ruan, Li, Liu, and Wu]{liu2025inference}
Zijun Liu, Peiyi Wang, Runxin Xu, Shirong Ma, Chong Ruan, Peng Li, Yang Liu, and Yu~Wu.
\newblock Inference-time scaling for generalist reward modeling.
\newblock \emph{arXiv preprint arXiv:2504.02495}, 2025.

\bibitem[Rein et~al.(2024)Rein, Hou, Stickland, Petty, Pang, Dirani, Michael, and Bowman]{rein2024gpqa}
David Rein, Betty~Li Hou, Asa~Cooper Stickland, Jackson Petty, Richard~Yuanzhe Pang, Julien Dirani, Julian Michael, and Samuel~R Bowman.
\newblock Gpqa: A graduate-level google-proof q\&a benchmark.
\newblock In \emph{First Conference on Language Modeling}, 2024.

\bibitem[Shiwen et~al.(2024)Shiwen, Liang, Liu, Zeng, and Liu]{skyworkcritic2024}
Tu~Shiwen, Zhao Liang, Chris~Yuhao Liu, Liang Zeng, and Yang Liu.
\newblock Skywork critic model series.
\newblock \url{https://huggingface.co/Skywork}, September 2024.
\newblock URL \url{https://huggingface.co/Skywork}.

\bibitem[Tan et~al.(2024)Tan, Zhuang, Montgomery, Tang, Cuadron, Wang, Popa, and Stoica]{tan2024judgebench}
Sijun Tan, Siyuan Zhuang, Kyle Montgomery, William~Y Tang, Alejandro Cuadron, Chenguang Wang, Raluca~Ada Popa, and Ion Stoica.
\newblock Judgebench: A benchmark for evaluating llm-based judges.
\newblock \emph{arXiv preprint arXiv:2410.12784}, 2024.

\bibitem[team(2025)]{qwen3}
Qwen team.
\newblock Qwen3: Think deeper, act faster.
\newblock \url{https://qwenlm.github.io/blog/qwen3/}, 2025.

\bibitem[Wang et~al.(2024)Wang, Ma, Zhang, Ni, Chandra, Guo, Ren, Arulraj, He, Jiang, et~al.]{wang2024mmlu}
Yubo Wang, Xueguang Ma, Ge~Zhang, Yuansheng Ni, Abhranil Chandra, Shiguang Guo, Weiming Ren, Aaran Arulraj, Xuan He, Ziyan Jiang, et~al.
\newblock Mmlu-pro: A more robust and challenging multi-task language understanding benchmark.
\newblock In \emph{The Thirty-eight Conference on Neural Information Processing Systems Datasets and Benchmarks Track}, 2024.

\bibitem[Yang et~al.(2024)Yang, Yang, Zhang, Hui, Zheng, Yu, Li, Liu, Huang, Wei, et~al.]{yang2024qwen2}
An~Yang, Baosong Yang, Beichen Zhang, Binyuan Hui, Bo~Zheng, Bowen Yu, Chengyuan Li, Dayiheng Liu, Fei Huang, Haoran Wei, et~al.
\newblock Qwen2. 5 technical report.
\newblock \emph{arXiv preprint arXiv:2412.15115}, 2024.

\bibitem[Ye et~al.(2024)Ye, Li, Li, Ai, Zhou, Shen, Yan, and Liu]{ye2024scalarrewardmodellearning}
Ziyi Ye, Xiangsheng Li, Qiuchi Li, Qingyao Ai, Yujia Zhou, Wei Shen, Dong Yan, and Yiqun Liu.
\newblock Beyond scalar reward model: Learning generative judge from preference data, 2024.
\newblock URL \url{https://arxiv.org/abs/2410.03742}.

\bibitem[Ye et~al.(2025)Ye, Li, Li, Ai, Zhou, Shen, Yan, and Liu]{ye2025learning}
Ziyi Ye, Xiangsheng Li, Qiuchi Li, Qingyao Ai, Yujia Zhou, Wei Shen, Dong Yan, and Yiqun Liu.
\newblock Learning llm-as-a-judge for preference alignment.
\newblock In \emph{The Thirteenth International Conference on Learning Representations}, 2025.

\bibitem[Yu et~al.(2025)Yu, Sun, Hu, Yan, Yu, and Li]{yu2025improve}
Jiachen Yu, Shaoning Sun, Xiaohui Hu, Jiaxu Yan, Kaidong Yu, and Xuelong Li.
\newblock Improve llm-as-a-judge ability as a general ability.
\newblock \emph{arXiv preprint arXiv:2502.11689}, 2025.

\bibitem[Zhou et~al.(2024)Zhou, Zheng, Wang, Xi, Dou, Bao, Shen, Xiong, Fan, Mou, et~al.]{zhou2024rmb}
Enyu Zhou, Guodong Zheng, Binghai Wang, Zhiheng Xi, Shihan Dou, Rong Bao, Wei Shen, Limao Xiong, Jessica Fan, Yurong Mou, et~al.
\newblock Rmb: Comprehensively benchmarking reward models in llm alignment.
\newblock \emph{arXiv preprint arXiv:2410.09893}, 2024.

\bibitem[Zhou et~al.(2023)Zhou, Lu, Mishra, Brahma, Basu, Luan, Zhou, and Hou]{zhou2023instruction}
Jeffrey Zhou, Tianjian Lu, Swaroop Mishra, Siddhartha Brahma, Sujoy Basu, Yi~Luan, Denny Zhou, and Le~Hou.
\newblock Instruction-following evaluation for large language models.
\newblock \emph{arXiv preprint arXiv:2311.07911}, 2023.

\bibitem[Zhu et~al.(2023)Zhu, Wang, and Wang]{zhu2023judgelm}
Lianghui Zhu, Xinggang Wang, and Xinlong Wang.
\newblock Judgelm: Fine-tuned large language models are scalable judges.
\newblock \emph{arXiv preprint arXiv:2310.17631}, 2023.

\end{thebibliography}

\clearpage
\appendix
\onecolumn
\section*{\Large{Appendix}}
\label{sec:Appendix}
\section{Deriving the Loss Function}
\para{Judge Reward.}
In the judge task, given a instruction-response pair $(x,y)$, prediction position $k_x$ and ground truth label $y_{k}^{*}$, we apply a rule-based reward defined as:

\begin{equation}
r ( x, y )=\begin{cases} 
1 & \text{if } y_{k_x}=y_{k_x}^{*} \\ 
0 & \text{otherwise} 
\end{cases}.
\end{equation}

\para{Policy Gradient Optimization.}
To optimize the judge model's performance, we formulate the learning objective as maximizing the expected reward over the response distribution:
\begin{equation}
J(\theta) = \mathbb{E}_{x \sim D} \mathbb{E}_{y \sim \pi_{\theta}} [r(x,y)]
\end{equation}
The gradient of this objective can be derived using the policy gradient theorem:
\begin{equation}
    \begin{aligned}
\nabla_{\theta} J(\theta) &= \mathbb{E}_{x \sim D} \mathbb{E}_{y \sim \pi_{\theta}} \left[ r(x,y) \nabla_{\theta} \log \pi_{\theta}(y|x) \right] \\
&= \mathbb{E}_{x \sim D} \mathbb{E}_{y \sim \pi_{\theta}} \left[ r(x,y) \sum_{t=1}^{n} \nabla_{\theta} \log \pi_{\theta}(y_t | x, y_{<t}) \right]
    \end{aligned}
\end{equation}
This decomposition shows how the gradient propagates through all sequence positions in autoregressive models.
The corresponding policy gradient loss is:
\begin{equation}
\mathcal{L}_{\mathrm{PG}} = -\mathbb{E}_{x \sim D} \mathbb{E}_{y \sim \pi_{\theta}} \left[ r(x,y) \sum_{t=1}^{n} \log \pi_{\theta}(y_t | x, y_{<t}) \right]
\end{equation}
Given our binary reward function that only depends on the prediction at position $k_x$, we can simplify:
\begin{equation}
    \label{eq:pg_loss}
\begin{aligned}
\mathcal{L}_{\mathrm{PG}} &= -\mathbb{E}_{x \sim D} \mathbb{E}_{y \sim \pi_{\theta}} \left[ r(x,y) \log \pi_{\theta}(y_{k_x} | x, y_{< {k_x}}) \right] \\
&= -\mathbb{E}_{x \sim D} \mathbb{E}_{y \sim \pi_{\theta}} \left[ \mathbb{I}(y_{k_x} = y_{k_x}^*) \log \pi_{\theta}(y_{k_x} | x, y_{< {k_x}}) \right] \\
&= -\mathbb{E}_{x \sim D} \mathbb{E}_{y \sim \pi_{\theta}} \left[ \log \pi_{\theta}(y_{k_x} | x, y_{< {k_x}} ) \Big | _{y_{k_x} = y_{k_x}^*} \right]
\end{aligned}
\end{equation}

\para{Rejection Sampling for RL Generalization.}
We further apply rejection sampling to approximate the policy gradient loss. 
Formally, for the $i_{th} $  instruction $(i \in \{1..,N\})$ in the dataset,  we generate $M$ response samples that satisfy the ground truth label $y^{(i,*)}$ and obtain the following loss:
\begin{equation}
    \begin{aligned}
L_{\mathrm{PG}} &= -\mathbb{E}_{x \sim D} \mathbb{E}_{y \sim \pi_{\theta}} \left[ \log \pi_{\theta}(y_{k_x} | x, y_{< {k_x}} ) \Big | _{y_{k_x} = y_{k_x}^*} \right] \\
&= -\frac{1}{N} \sum_{i=1}^{N} \frac{1}{M} \sum_{j=1}^{M} \left[ \log \pi_{\theta}(y_{k_x}^{(i,j)} | x^{(i)}, y_{< {k_x}}^{(i,j)} ) \Big | _{y_{k_x}^{(i,j)} = y_{k_x}^{(i,*)}} \right] \\
&= -\frac{1}{NM} \sum_{i=1}^{N} \sum_{j=1}^{M} \log \pi_{\theta} ( y_{k_x}^{(i,*)} | x^{(i)}, y_{< k_x}^{(i,*)} )
\end{aligned}
\end{equation}

Similarly, we apply the SFT loss to the sampled response candidates. To balance the standard sequence modeling objective with reward optimization, we combine the SFT loss and policy gradient loss through a mapping function $f$ and derive another mapping function $g$:

\begin{equation}
    \begin{aligned}
    \mathcal{L} &= \mathcal{L}_{\mathrm{SFT}} + \mathcal{L}_{\mathrm{PG}}(f) \\
        &= -\frac{1}{NM} \sum_{i=1}^{N}\sum_{j=1}^{M} \sum_{t \neq k_{ij}} \log \pi_{\theta} ( y_{t}^{(i,j)} | x^{(j)}, y_{< t}^{(i,j)} ) \\
        &\quad -\frac{1}{NM} \sum_{i=1}^{N}\sum_{j=1}^{M} \log \pi_{\theta} ( y_{k_{ij}}^{(i,*)} | x^{(j)}, y_{< k_{ij}}^{(i,j)} )  + f\left(\log \pi_{\theta} ( y_{k_{ij}}^{(i,*)} | x^{(j)}, y_{< k_{ij}}^{(i,j)} )\right) \\
        &= -\frac{1}{NM} \sum_{i=1}^{N}\sum_{j=1}^{M}  \left[ \sum_{t \neq k_{ij}} \log \pi_{\theta} ( y_{t}^{(i,j)} | x^{(j)}, y_{< t}^{(i,j)} ) +  g\left(\log \pi_{\theta} ( y_{k_{ij}}^{(i,*)} | x^{(j)}, y_{< k_{ij}}^{(i,j)} )\right) \right]
    \end{aligned}, 
\end{equation}
where $g$ is a composite function that combines the original mapping function $f$ with the log probability term to provide a more flexible optimization objective. In our method, the mapping function $g$ is approximate by DPO loss, Temperture Loss and Margin Loss.

\section{Detailed Results on the Judge Benchmarks}
We list the detailed results of judge models on the Judge Benchmarks in Table \ref{tab:detail_judgerbench}, \ref{tab:detail_rmb} and \ref{tab:detail_rewardbench}. 
\begin{table*}[!h]
    \centering
    \caption{\textbf{Detailed results on JudgerBenchV2 benchmarks.} }
    \label{tab:detail_judgerbench}
    \resizebox{.9\textwidth}{!}{
    \begin{tabular}{lccccc}
    \toprule
    \textbf{Model} &\textbf{Accuracy} & \textbf{Normalized Diff} & \textbf{Rank Diff} & \textbf{Score Diff} & \textbf{Final Score} \\      
    \midrule
    \multicolumn{6}{c}{\textbf{7B Judge Models}} \\ \midrule
    CompassJudger-1-7B-Instruct \cite{cao2024compassjudger} & 77.41 & 61.48 & 11.40 & 83.40 & 57.96 \\
    Con-J-Qwen2-7B \cite{ye2025learning} & 71.30 & 66.61 & 17.60 & 85.20 & 52.35 \\
    RISE-Judge-Qwen2.5-7B \cite{yu2025improve} & 70.08 & 77.85 & 14.00 & 202.50 & 46.12 \\
    \rowcolor{tabhighlight} \model{7B} & 78.04 & 57.00 & 10.80 & 76.90 & 60.52 \\
    \midrule
    \multicolumn{6}{c}{\textbf{32B+ Judge Models}} \\ \midrule
    CompassJudger-1-32B-Instruct \cite{cao2024compassjudger} & 80.99 & 60.32 & 11.40 & 62.90 & 60.33 \\
    Skywork-Critic-Llama-3.1-70B \cite{skyworkcritic2024} & 70.27 & 65.44 & 15.20 & 97.30 & 52.41 \\
    RISE-Judge-Qwen2.5-32B \cite{yu2025improve} & 74.00 & 61.15 & 10.60 & 88.80 & 54.42 \\
    \rowcolor{tabhighlight} \model{32B} & 80.90 & 56.47 & 8.60 & 64.10 & 62.21 \\
    \bottomrule
    \end{tabular}
    }
\end{table*}

\begin{table*}[!h]
    \centering
    \caption{\textbf{Detailed results on RMB benchmarks.} }
    \label{tab:detail_rmb}
    \resizebox{.9\textwidth}{!}{
    \begin{tabular}{lccc}
    \toprule
    \textbf{Model} &\textbf{Pair Accuracy} & \textbf{BoN Accuracy} & \textbf{Final Score} \\      
    \midrule
    \multicolumn{4}{c}{\textbf{7B Judge Models}} \\ \midrule
    CompassJudger-1-7B-Instruct \cite{cao2024compassjudger} & 47.40 & 28.96 & 38.18 \\
    Con-J-Qwen2-7B \cite{ye2025learning}  & 84.80 & 74.20 & 79.50 \\
    RISE-Judge-Qwen2.5-7B \cite{yu2025improve} & 78.79 & 66.50 & 72.64 \\
    \rowcolor{tabhighlight} \model{7B} & 80.58 & 67.23 & 73.90 \\
    \midrule
    \multicolumn{4}{c}{\textbf{32B+ Judge Models}} \\ \midrule
    CompassJudger-1-32B-Instruct \cite{cao2024compassjudger} & 82.73 & 72.53 & 77.63 \\
    Skywork-Critic-Llama-3.1-70B \cite{skyworkcritic2024} & 68.35 & 62.50 & 65.50 \\
    RISE-Judge-Qwen2.5-32B \cite{yu2025improve} & 79.99 & 67.42 & 73.70 \\
    \rowcolor{tabhighlight} \model{32B} & 79.61 & 66.35 & 72.98 \\
    \bottomrule
    \end{tabular}
    }
\end{table*}

\begin{table*}[!h]
    \centering
    \caption{\textbf{Detailed results on RewardBench benchmarks.} }
    \label{tab:detail_rewardbench}
    \resizebox{.9\textwidth}{!}{
    \begin{tabular}{lccccc}
    \toprule
    \textbf{Model} & \textbf{Chat} & \textbf{Chat Hard} & \textbf{Safety} & \textbf{Reasoning} & \textbf{Final Score} \\      
    \midrule
    \multicolumn{6}{c}{\textbf{7B Judge Models}} \\ \midrule
    CompassJudger-1-7B-Instruct \cite{cao2024compassjudger} & 97.80 & 61.00 & 84.50 & 89.50 & 83.20 \\
    Con-J-Qwen2-7B \cite{ye2025learning} & 91.90 & 80.30 & 88.20 & 88.10 & 87.10 \\
    RISE-Judge-Qwen2.5-7B \cite{yu2025improve} & 92.20 & 76.50 & 88.00 & 96.10 & 88.20 \\
    \rowcolor{tabhighlight} \model{7B} & 92.36 & 85.99 & 91.08 & 94.41 & 90.96 \\
    \midrule
    \multicolumn{6}{c}{\textbf{32B+ Judge Models}} \\ \midrule
    CompassJudger-1-32B-Instruct \cite{cao2024compassjudger} & 98.00 & 65.10 & 85.30 & 92.40 & 85.20 \\
    Skywork-Critic-Llama-3.1-70B \cite{skyworkcritic2024} & 96.60 & 87.90 & 93.10 & 95.50 & 93.30 \\
    RISE-Judge-Qwen2.5-32B \cite{yu2025improve} & 96.60 & 83.30 & 91.90 & 98.80 & 92.70 \\
    \rowcolor{tabhighlight} \model{32B} & 93.37 & 88.58 & 90.68 & 97.00 & 92.40 \\
    \bottomrule
    \end{tabular}
    }
\end{table*}

\newpage

\begin{figure*}
\begin{tcolorbox}[
colback=white!10!white,
colframe=black!50!white,
title=CoT Synthesizing Prompt,
]
Now we are reviewing a user's interaction with two models. Your task is to evaluate the responses from Model A and Model B by carefully analyzing the dialogue step by step, following a clear and structured thought process:

1. User's Demand:

   - Carefully analyze the user's request. What is the user specifically asking for? What are the key aspects of the request that need to be fulfilled? Identify any constraints (e.g., time, format, quantity) the user has provided.

2. Strengths of Model A:

   - Identify the strengths of Model A's response. Consider how well it addresses the user's demand, meets the user’s constraints, and how well it serves the intended purpose.

3. Weaknesses of Model A:

   - Identify the weaknesses of Model A's response. What aspects of the response fail to meet the user's request or constraints? What could have been improved?

4. Strengths of Model B:

   - Identify the strengths of Model B's response. Consider how well it addresses the user's demand, meets the user’s constraints, and how well it serves the intended purpose.

5. Weaknesses of Model B:

   - Identify the weaknesses of Model B's response. What aspects of the response fail to meet the user's request or constraints? What could have been improved?

6. Reasoning:

   - Based on your analysis of both responses, explain which model better addresses the user's needs. Discuss which model's response is more suitable given the user's request and constraints.

7. Choice:

   - Conclude with a choice between Model A and Model B based on your reasoning. Indicate which model provides the more appropriate and useful response for the user's request.

Your final reply must be structured in the following format:

\{

  "User's Demand": "[The user's request or need]",

  "Strengths of Model A": "[Summary of the strengths of Model A]",

  "Weaknesses of Model A": "[Summary of the weaknesses of Model A]",

  "Strengths of Model B": "[Summary of the strengths of Model B]",

  "Weaknesses of Model B": "[Summary of the weaknesses of Model B]",

  "Reasoning": "[Explanation of which model is more suitable for the user's demand]",

  "Choice": "[Model A or Model B]"

\}

\label{fig:critique-prompt}
\end{tcolorbox}
\end{figure*}

\begin{figure*}
\begin{tcolorbox}[
colback=white!10!white,
colframe=black!50!white,
title=Critique Example-Prompt,
]

Below is a user's question and a model's response. Please help identify the strengths and weaknesses of the model's reply so that it can better address the user's question.  

The user's question is as follows:  
Write a poem about winter, requiring each sentence to be exactly five words long, and without using the words "winter" or "snow," yet still evoking a sense of the season.

The model's response is as follows:  
Frost paints silent trees.  Chill lingers in the air.  Footsteps fade on paths.  Windows gleam with light.  Hushed, the world awaits.  

Now, please provide your feedback, highlighting the strengths and areas for improvement in the model's response.

\label{fig:critique-prompt}
\end{tcolorbox}
\end{figure*}

\begin{figure*}
\begin{tcolorbox}[
colback=white!10!white,
colframe=black!50!white,
title=Critique Example-RISE-Judge-Qwen2.5-7B Answer,
]

Strengths of the Model's Response:

1. Conciseness and Adherence to Word Count: Each sentence is exactly five words long, as requested by the user.

2. Evocation of Winter: The response effectively evokes the sense of winter without explicitly using the words "winter" or "snow." Phrases like "Frost paints silent trees" and "Chill lingers in the air" convey the cold and serene aspects of winter.

3. Imagery and Atmosphere: The response uses vivid imagery to create a strong sense of atmosphere. Descriptions such as "Footsteps fade on paths" and "Windows gleam with light" paint a clear picture of a winter scene.

4. Coherence and Flow: The sentences flow well together, creating a coherent and engaging poem.

Areas for Improvement:

\red{1.Inclusion of "Snow": While the user specifically asked to avoid the word "snow," it is a common and essential element of winter that could be subtly included to enhance the evocation of the season. For example, "Frost paints silent trees" could be modified to "Snow blankets silent trees" to strengthen the winter theme.}

2. Variety in Sentence Structure: The response has a consistent structure with each sentence being a simple statement. Introducing a bit of variation in sentence structure could add to the poetic quality, such as using a question or exclamation to create a different tone or emphasis.

3. Enhancing Emotional Depth: While the response is atmospheric, adding a bit more emotional depth or personal touch could make the poem more engaging. For example, describing the "hushed, waiting world" as "yearning for spring" could add a layer of emotional resonance.

Overall, the model's response is strong and meets most of the user's requirements effectively. However, subtle adjustments could enhance its impact and completeness.

\label{fig:rise-answer}
\end{tcolorbox}
\end{figure*}

\begin{figure*}
\begin{tcolorbox}[
colback=white!10!white,
colframe=black!50!white,
title=Critique Example-CompassJudger-1-7B-Instruct Answer,
]

Strengths:

1. The model adheres to the instruction of creating a poem where each sentence is exactly five words long.

2. The poem effectively evokes the essence of winter without explicitly using the words "winter" or "snow."

3. The language used is descriptive and creates a vivid image of the season, such as "Frost paints silent trees" and "Windows gleam with light."

4. There is a coherent flow and structure to the poem, with each line contributing to a unified theme of winter.

Areas for Improvement:

1. While the poem avoids the word "snow," which was part of the instruction, \red{the absence of any snow-related imagery} or metaphor might have strengthened the winter theme further.

2. The phrase "Hushed, the world awaits" could be interpreted as a generic description of anticipation and not necessarily tied to the cold or silence of winter, which might make the winter theme slightly less explicit.

3. To further enhance the specificity of the season, incorporating additional sensory details or specific winter activities (like ice skating or \red{building snowmen}) could make the poem more evocative and engaging.

Overall, the model's response is a well-crafted poem that meets most of the criteria set by the user's request. With a few minor adjustments, it could become even more effective in capturing the spirit of winter.

\label{fig:cj1-answer}
\end{tcolorbox}
\end{figure*}

\begin{figure*}
\begin{tcolorbox}[
colback=white!10!white,
colframe=black!50!white,
title=Critique Example-RISE-Judge-Qwen2.5-7B Answer,
]

Strengths of the Model's Response:

1. Conciseness and Adherence to Word Count: Each sentence is exactly five words long, as requested by the user.

2. Evocation of Winter: The response effectively evokes the sense of winter without explicitly using the words "winter" or "snow." Phrases like "Frost paints silent trees" and "Chill lingers in the air" convey the cold and serene aspects of winter.

3. Imagery and Atmosphere: The response uses vivid imagery to create a strong sense of atmosphere. Descriptions such as "Footsteps fade on paths" and "Windows gleam with light" paint a clear picture of a winter scene.

4. Coherence and Flow: The sentences flow well together, creating a coherent and engaging poem.

Areas for Improvement:

\red{1.Inclusion of "Snow": While the user specifically asked to avoid the word "snow," it is a common and essential element of winter that could be subtly included to enhance the evocation of the season. For example, "Frost paints silent trees" could be modified to "Snow blankets silent trees" to strengthen the winter theme.}

2. Variety in Sentence Structure: The response has a consistent structure with each sentence being a simple statement. Introducing a bit of variation in sentence structure could add to the poetic quality, such as using a question or exclamation to create a different tone or emphasis.

3. Enhancing Emotional Depth: While the response is atmospheric, adding a bit more emotional depth or personal touch could make the poem more engaging. For example, describing the "hushed, waiting world" as "yearning for spring" could add a layer of emotional resonance.

Overall, the model's response is strong and meets most of the user's requirements effectively. However, subtle adjustments could enhance its impact and completeness.

\label{fig:rise-answer}
\end{tcolorbox}
\end{figure*}

\begin{figure*}
\begin{tcolorbox}[
colback=white!10!white,
colframe=black!50!white,
title=Critique Example-CompassJudger-2-7B-Instruct Answer,
]

Strengths:

1. Adherence to Constraints: The model successfully adheres to the requirement of each sentence being exactly five words long and avoids using the words "winter" or "snow."

2. Imagery and Atmosphere: The poem effectively evokes a sense of the season through vivid imagery and subtle hints. For example, "Frost paints silent trees" and "Chill lingers in the air" convey the cold and stillness characteristic of winter.

3. Consistency in Tone: The tone remains consistent throughout the poem, maintaining a sense of hushed and serene atmosphere that aligns with the winter season.

Areas for Improvement:

1. Variety in Imagery: While the imagery is strong, it could benefit from a bit more variety. For example, including different sensory details (e.g., sounds, smells) could make the poem more immersive.

2. Emotional Depth: Adding a slight emotional undertone could enhance the poem. For instance, a line that hints at nostalgia or anticipation could deepen the reader's connection to the season.

3. Rhythm and Flow: The rhythm is good, but slightly varying the structure or adding a rhyming scheme could make the poem more engaging and memorable.

\label{fig:cj2-answer}
\end{tcolorbox}
\end{figure*}
\end{document}